\documentclass[runningheads]{llncs}

\usepackage{eccv}

\usepackage[dvipsnames]{xcolor}

\usepackage{amsmath}
\usepackage{amsfonts}
\usepackage{amssymb}
\usepackage{algorithmic}
\usepackage{algorithm}
\usepackage{array}

\usepackage{textcomp}
\usepackage{verbatim}
\usepackage{graphicx}
\usepackage{cite}

\usepackage{xspace}

\usepackage{caption}

\usepackage{wrapfig}

\usepackage{booktabs}
\usepackage{threeparttable}
\usepackage{enumitem}
\usepackage{colortbl}
\usepackage{multirow}
\usepackage{stfloats}

\usepackage{xcolor}

\usepackage{bm}
\usepackage{siunitx}

\usepackage{adjustbox}
\usepackage{pifont}

\definecolor{green1}{HTML}{008000}
\definecolor{green2}{HTML}{75b975}
\definecolor{green3}{HTML}{ebf3eb}
\usepackage{eccvabbrv}

\usepackage{graphicx}
\usepackage{booktabs}

\usepackage[accsupp]{axessibility}  % Improves PDF readability for those with disabilities.

\usepackage{hyperref}

\usepackage{orcidlink}

\begin{document}

% \title{Author Guidelines for ECCV Submission}
\title{On the Evaluation Consistency of Attribution-based Explanations}

\makeatletter
\newcommand{\myfnsymbol}[1]{%
  \expandafter\@myfnsymbol\csname c@#1\endcsname
}
% Mapping of how the \thanks symbols will be interpreted sequentially
\newcommand{\@myfnsymbol}[1]{%
  \ifcase #1
    % 0
  \or \TextOrMath{\textasteriskcentered}{*}% 1
  \or \TextOrMath{\textdagger}{\dagger}% 2
  \fi
}
\renewcommand{\thefootnote}{\myfnsymbol{footnote}}

\makeatother

\author{Jiarui Duan\inst{1,3} \and
Haoling Li\inst{2,3} \and
Haofei Zhang\inst{1,3} \and
Hao Jiang\inst{4} \and
Mengqi Xue\inst{5} \and
\\
Li Sun\inst{6} \and
Mingli Song\inst{1,3} \and
Jie Song\inst{2,3}\textsuperscript{$\dagger$}
}

\authorrunning{J.~Duan et al.}

% Springer Heidelberg, Tiergartenstr.~17, 69121 Heidelberg, Germany
\institute{
College of Computer Science and Technology, Zhejiang University \and
School of Software Technology, Zhejiang University \and
Key Laboratory of Visual Perception (Zhejiang University), Ministry of Education and Microsoft \and
Alibaba Group \and
Hangzhou City University \and
Ningbo Innovation Center, Zhejiang University\\
\email{\{jerryduan,hollylee,haofeizhang,lsun,brooksong,sjie\}@zju.edu.cn};
\email{aoshu.jh@alibaba-inc.com};
\email{mqxue@hzcu.edu.cn}
}

\maketitle
\footnotetext[2]{Corresponding author.}
\setcounter{footnote}{0}
\renewcommand{\thefootnote}{\arabic{footnote}}

\begin{abstract}
  Attribution-based explanations are garnering increasing attention recently and have emerged as the predominant approach towards \textit{eXplanable Artificial Intelligence}~(XAI). However, the absence of consistent configurations and systematic investigations in prior literature impedes comprehensive evaluations of existing methodologies. In this work, we introduce {Meta-Rank}, an open platform for benchmarking attribution methods in the image domain.  Presently, Meta-Rank assesses eight exemplary attribution methods using six renowned model architectures on four diverse datasets, employing both the \textit{Most Relevant First} (MoRF) and \textit{Least Relevant First} (LeRF) evaluation protocols.
  Through extensive experimentation, our benchmark reveals three insights in attribution evaluation endeavors:
  1) evaluating attribution methods under disparate settings can yield divergent performance rankings; 2) although inconsistent across numerous cases, the performance rankings exhibit remarkable consistency across distinct checkpoints along the same training trajectory; 3) prior attempts at consistent evaluation fare no better than baselines when extended to more heterogeneous models and datasets. Our findings underscore the necessity for future research in this domain to conduct rigorous evaluations encompassing a broader range of models and datasets, and to reassess the assumptions underlying the empirical success of different attribution methods.
  % Code, models, and datasets will be made publicly available in the near future.
  Our code is publicly available at \url{https://github.com/TreeThree-R/Meta-Rank}.
\end{abstract}

\section{Introduction}
\label{sec:intro}
Explainable Artificial Intelligence (XAI)~\cite{adadi2018peeking} has emerged as a prominent research field within computer vision, with a multitude of approaches proposed for explaining and interpreting the internal mechanisms of diverse neural network architectures across various application domains. Among these approaches, attribution methods have gained significant traction in the realm of deep learning. By generating an explanation map in the form of a heatmap~(also referred to as a relevance map or a saliency map), these methods attribute the prediction to specific regions within the input image, thereby facilitating interpretation. Over the years, researchers have devised a large number of attribution methods from diverse perspectives, including gradient-based methods~\cite{shrikumar2016not, sundararajan2017axiomatic, springenberg2014striving, shrikumar2017learning, bach2015pixel}, perturbation-based methods~\cite{petsiuk2018rise, zeiler2014visualizing, lundberg2017unified, ribeiro2016should}, and CAM-based methods~\cite{selvaraju2017grad, chattopadhay2018grad}.

Nevertheless, given the absence of ground truth regarding the inner workings of deep neural networks, evaluating attributions directly poses a formidable challenge. Presently, various studies have endeavored to assess the reliability of attribution methods, which can be roughly categorized into two schools: \textit{expert-grounded} methods~\cite{zhang2018top,nguyen2021effectiveness,bastings2019interpretable,covert2020understanding,colin2022cannot,krishna2022disagreement}, and \textit{functional-grounded} methods~\cite{samek2016evaluating, petsiuk2018rise, hooker2019benchmark, rong2022consistent, deng2023understanding, atanasova2020diagnostic,wang2022unified}.
Expert-grounded evaluations capture how well the explanations imitate the human-annotated importance of the image regions. Such methods not only require human effort, but also suffer from the faithfulness\footnotemark~issue, as deep neural networks could rely on non-intuitive, spurious cues that differ from human perception~\cite{geirhos2020shortcut}.
In contrast, functional-based approaches evaluate attribution methods based on some fundamental axioms.
For instance, feature ablation~\cite{petsiuk2018rise}, as a functional-grounded evaluation, systematically removes features in a predetermined order and observes the resulting changes in predictions. This approach enables more faithful assessments of the attribution explanations by comparing the model behavior with and without specific individual features~\cite{ribeiro2016should, petsiuk2018rise, hooker2019benchmark, fong2017interpretable, zintgraf2017visualizing, goyal2019counterfactual}.
Unfortunately, the current literature often evaluates attributions under divergent settings, with different datasets, models and evaluation protocols~\cite{hooker2019benchmark,rong2022consistent}. The efficacy of attribution methods in practice is often called into question due to restricted and inconsistent experimental conditions.

\footnotetext{Faithfulness refers to the degree of alignment between the results intepreted by an attribution method and the decision of the model.}

The goal of this work is to gain a thorough understanding of
the current state of attribution methods while setting the stage for critical problems to be worked on. To this end, we present
Meta-Rank, an open-sourced attribution benchmark featuring rigorous evaluations, comprehensive analyses as well as extensive baselines. Our benchmark carefully examines eight widely-used attribution methods~(Saliency~\cite{simonyan2014deep}, Input$\odot$Gradient~\cite{shrikumar2016not}, Integrated Gradients~\cite{sundararajan2017axiomatic}, Guided Backpropagation~\cite{springenberg2014striving}, DeepLift~\cite{shrikumar2017learning}, Deconvolution~\cite{zeiler2014visualizing}, LRP~\cite{bach2015pixel} and Guided Grad-CAM~\cite{selvaraju2017grad}), with a wide range of models~(ResNet-18~\cite{he2016deep}, ResNet-50~\cite{he2016deep}, Inception-v4~\cite{szegedy2017inception}, VGG-19~\cite{simonyan2014very}, EfficientNet\_b2~\cite{tan2019efficientnet}, and DenseNet-121~\cite{huang2017densely}), on various domain datasets~(NWPU-RESISC45~\cite{cheng2017remote}, Food-101~\cite{bossard2014food}, ImageNet-1k~\cite{deng2009imagenet}, and Places-365~\cite{zhou2017places}), using two evaluation protocols~(\textit{Most Relevant First} and \textit{Least Relevant First}). Specifically, we place a strong emphasis on comprehensive and consistent experimental settings that have been largely overlooked in previous works. Through extensive experiments on the exhaustive combinations of methods, datasets, models and evaluation protocols, our benchmark reveals three main findings regarding previous attribution evaluation endeavors:
1) evaluating attribution methods under disparate experimental settings, such as different model architectures, datasets and evaluation protocols, can yield highly divergent performance rankings;
2) although highly inconsistent across datasets, models and evaluation protocols, the performance rankings exhibit remarkable consistency across distinct checkpoints along the same training trajectory;
3) prior attempts towards more consistent evaluation, like Remove and Debias~(ROAD)~\cite{rong2022consistent}, fare no better than simple baselines when it is extended to more heterogeneous models and datasets.

We further propose a benchmark metric, named Meta-Rank, to facilitate a more comprehensive benchmark comparison between attribution methods. Meta-Rank aggregates ranking results from a diverse array of settings, thereby mitigating biases associated with specific configurations. By employing this metric, we conduct systematic comparisons among existing attribution methods, and provide extensive experiment results and discussions on the status quo. Beyond extensive experiments, our benchmark is designed as a flexible framework that standardizes experimental settings and simplifies the integration of novel algorithmic implementations.
We hope our work will not only facilitate reliable evaluations of attribution methods across a broader range of models and datasets, but also inspire the research community to delve further into the development of more consistent attribution methods.

In summary, we make the following contributions:
\begin{itemize}[label=\textbullet]
    \item We present Meta-Rank, an open-sourced attribution benchmark featuring rigorous evaluations, comprehensive analyses as well as extensive baselines.
    \item Through extensive experiments, we identify three pivotal insights from previous attribution evaluation efforts, which provide directions for future advancements in attribution research.
    \item We propose an evaluation metric to facilitate a more comprehensive benchmark comparison across multiple settings, with which extensive experiments and discussions are provided on the status quo.
\end{itemize}

\section{Related Work}
The development of attribution methods has prompted researchers to seek the best approach to assist models in downstream tasks.
In this section, we delineate the progress of attribution evaluation chronologically.

\noindent \textbf{Expert-grounded evaluations}~\cite{zhang2021fine, zhang2018top, nguyen2021effectiveness, bastings2019interpretable, covert2020understanding} rely on human judgment to evaluate attribution methods.
For example, Zhang \etal~\cite{zhang2018top} proposed the pointing-game method as a means of evaluating attention maps based on detecting whether the highest value point in the map hit the target category. However, such methods are inherently subjective and fail to guarantee the reliability of evaluation results.

\noindent \textbf{Functional-grounded evaluations}~\cite{ancona2017towards,petsiuk2018rise,samek2016evaluating,hooker2019benchmark,bhatt2020evaluating} provide more objective evaluations of the model by comparing how it behaves when certain individual features are present or absent.
Samek \etal~\cite{samek2016evaluating} proposed two evaluation protocols, namely \textit{Most Relevant First}~(MoRF) and \textit{Least Relevant First}~(LeRF), to observe the impact on the model's output. In MoRF, effective attribution methods should produce a greater reduction in model accuracy when removing features with higher attribution values, and vice verse.
Similarly, \cite{petsiuk2018rise} suggested evaluating attribution methods through the \textit{deletion} and \textit{insertion} of features by using their attribution values. These methods violate the assumption of a consistent distribution between training and test data, making it hard to determine if the decline in model performance solely results from feature deletion.
Hooker \etal~\cite{hooker2019benchmark} further demonstrated that the performance degradation from feature deletion partly stems from such a distribution shift. To address this, they developed \textit{RemOve And Retrain}~(ROAR), which resolves inconsistent distributions by synchronously deleting the features of the training set and test set based on the attributions of each attribution method, and then retrained the model on the generated datasets. The effectiveness of the attribution method is measured by the retrained model's accuracy decline. Better methods identify more important features, whose removal causes sharper performance drops.
ROAR has been widely adopted in evaluation studies~\cite{bhatt2020evaluating,ismail2020benchmarking,hartley2020explaining}.
However, due to its exorbitant expense, this type of evaluation method is unfeasible for scaling up.
Rong \etal~\cite{rong2022consistent} discovered inconsistent MoRF and LeRF rankings from ROAR, which they attributed to information leakage based on information theory analysis. Thus, they proposed a noisy linear imputation method, named \textit{RemOve And Debias}~(ROAD), to effectively generate consistent results. However, as a pixel filling method, ROAD still suffers from several inherent defects, \eg, the features are replaced with substantial values instead of entirely dropped, which can hardly guarantee their neutral effects on the predictions.
Deng \etal~\cite{deng2023understanding} noted the lack of unified theoretical frameworks for evaluating attribution methods. To resolve this limitation, they proposed a Taylor interaction system to unify fourteen attribution methods under a common mathematical formulation and suggested three principles for evaluation. However, applying this framework to new methods requires manual theoretical analysis, hampering both scalability and efficiency.
Recently, attribution evaluation has been explored from more diverse perspectives~\cite{zhou2022feature, khakzar2022explanations, rao2023better}, including comprehensive viewpoints~\cite{hedstrom2023quantus,hesse2023funnybirds} as well as the consistency lens~\cite{brunke2020evaluating,rong2022consistent,binder2023shortcomings,hooker2019benchmark}.
Unfortunately, the absence of standardized experimental settings and systematic evaluations in previous literatures poses challenges for comparing existing methods universally. To overcome this limitation, we propose Meta-Rank, a novel evaluation benchmark, to provide a comprehensive evaluation of attribution methods across multiple settings.

\section{Preliminaries}
\label{sec:pre}
Attribution methods explain deep models by generating saliency maps that highlight pixels critical for predictions.
In our work, we benchmark eight prevalent attribution methods: Saliency~\cite{simonyan2014deep}, Input$\odot$Gradient~\cite{shrikumar2016not}, Integrated Gradients~\cite{sundararajan2017axiomatic}, Guided Backpropagation~\cite{springenberg2014striving}, DeepLift~\cite{shrikumar2017learning}, Deconvolution~\cite{zeiler2014visualizing}, LRP~\cite{bach2015pixel}, and Guided Grad-CAM~\cite{selvaraju2017grad}.
More details regarding these methods are recorded in~Appendix~B.
In the following, we formalize the key concepts in attribution evaluation, deriving a general assessment paradigm in~XAI.

To begin with, let $\mathcal{D}=\{(\mathbf{x}_i, y_i)\}_{i=1}^n$ be an image classification dataset with $n$ samples and $C$ categories, where $y_i$ is the label of RGB image $\mathbf{x}_i \in \mathbb{R}^{3\times H\times W}$.
Considering a deep network $\hat{y} = f_{\Theta}(\mathbf{x})$ that predicts $\hat{y}$ for input $\mathbf{x}$, an attribution method aims to generate a heatmap $H\in\mathbb{R}^{3\times H\times W}$ so that each input pixel is assigned to a saliency score to the network's prediction.
Such a procedure can be formulated as the following form:
\begin{equation}
    H = \mathtt{Attr}_{\Gamma}(f_\Theta | \mathbf{x}, \hat{y}) \text{,}
\end{equation}
where $\mathtt{Attr}_{\Gamma}$ denotes a certain attribution method parameterized by $\Gamma$.

To evaluate the faithfulness of different attribution methods, we adopt two most common strategies: with progressively removing the most or least salient pixels of each image in the query dataset, we measure the performance degradation of the target model.
The two strategies are usually referred to as the \textit{Most Relevant First}~(MoRF) and the \textit{Least Relevant First}~(LeRF) evaluation protocols.
Specifically, we construct a top-$t$ removed query dataset $\tilde{\mathcal{D}}^{(t)}$ by removing the top-$t$ most or least salient pixels of the original image $\mathbf{x}_i$ and derive the modified image $\tilde{\mathbf{x}}^{(t)}_i$ as follows:
\begin{equation}
\tilde{\mathbf{x}}^{(t)}_i = \mathtt{Remove}(\mathbf{x}_i, H_i | t) \text{.}
\end{equation}
Here, the operation $\mathtt{Remove}$ specifies how features are eliminated, representing feature ablation in this paper.

Effective attribution methods can accurately pinpoint significant pixels, and the elimination of these pixels would result in a swift deterioration in the model's performance.
Thus, the evaluation of attribution methods can be facilitated by quantifying the reduction in model performance, which is defined by:
\begin{equation}
\Delta \mathbf{} = \frac{1}{n} (\sum\nolimits_{i=1}^{n} p(\hat{y}_i=y_i|\mathbf{x}_i) - \sum\nolimits_{i=1}^{n} p(\tilde{y}^{(t)}_i=y_i|\tilde{\mathbf{x}}^{(t)}_i)) \textit{.}
\end{equation}
In this work, we focus on faithfulness to acquire a broadly applicable standard among multiple experimental settings. Ultimately, attribution methods are ranked according to this metric to generate a ranking list.
The schematic diagram of the attribution and evaluation is depicted in Figure~\ref{fig:Meta-Rank}~(b).

\begin{table*}[!t]
    \tiny%
    \centering%
    \caption{
    Comparison of experimental configurations employed in previous studies. The lack of consistency in these settings may lead to disparate observations. The term ``Cons.'' refers to whether the study addresses the issue of consistency. Only a limited number of works have explored the issue, but their scope is limited to specific datasets, models, or both.
    }
    \begin{tabular}{ccrrc}
    \toprule
	\textbf{Methods} & \textbf{Venue} & \textbf{Datasets} & \textbf{Models} & \textbf{Cons.} \\
    \midrule
    \cite{simonyan2014deep} & ICLR 2014 & ImageNet & ConvNet & \ding{55} \\
    \cite{shrikumar2016not} & ICML 2016 & Tiny ImageNet, DNA sequence & VGG-16, 1 CNN  & \ding{55} \\
    \cite{sundararajan2017axiomatic} & ICML 2017 & ImageNet & GoogleNet & \ding{55} \\
    \cite{springenberg2014striving} & ICLR Workshop 2015 & ImageNet, CIFAR-10/100 & 3 CNNs & \ding{55} \\
    \cite{shrikumar2017learning} & ICML 2017 & MNIST, DNA sequence & 2 CNNs & \ding{55} \\
    \cite{zeiler2014visualizing} & ECCV 2014 & ImageNet, Caltech, PASCAL VOC & 2 CNNs & \ding{55} \\
    \cite{bach2015pixel} & PLOS ONE 2015 & ImageNet, PASCAL VOC, MINIST & 1 DNN, 1 CNN & \ding{55} \\
    \cite{selvaraju2017grad} & ICCV 2017 & ImageNet, PASCAL VOC & VGG-16, AlexNet & \ding{55} \\
	\cite{hooker2019benchmark} & NeurIPS 2019 & ImageNet, Birdsnap, Food-101 & ResNet-50 & \ding{55} \\
    \cite{brunke2020evaluating} & ECCV Workshop 2020 & ImageNet & ResNet-50 & \ding{51} \\
    \cite{arias2022focus} & FUZZ-IEEE 2022 & ImageNet, Dogs vs. Cats, MAMe, MIT67 & AlexNet, VGG-16, ResNet-18 & \ding{55} \\
	\cite{rong2022consistent} & ICML 2022 & CIFAR-10, Food-101 & ResNet-18, ResNet-50 & \ding{51} \\
    \cite{brocki2022evaluation} & ICLR Tiny Paper 2023 & ImageNet & ResNet-50 & \ding{55} \\
    \cite{binder2023shortcomings} & CVPR 2023 & ImageNet & VGG-16, ResNet-34 & \ding{51} \\
    \cite{hedstrom2023quantus} & JMLR 2023 & ImageNet & ResNet-18 & \ding{55} \\
    \bottomrule
    \end{tabular}
    \label{tab:settings}
\end{table*}

\section{Meta-Rank Settings and Benchmark}
\label{sec:meta}
\begin{figure*}[!t]
\centering%
\includegraphics[width=1.0\linewidth]{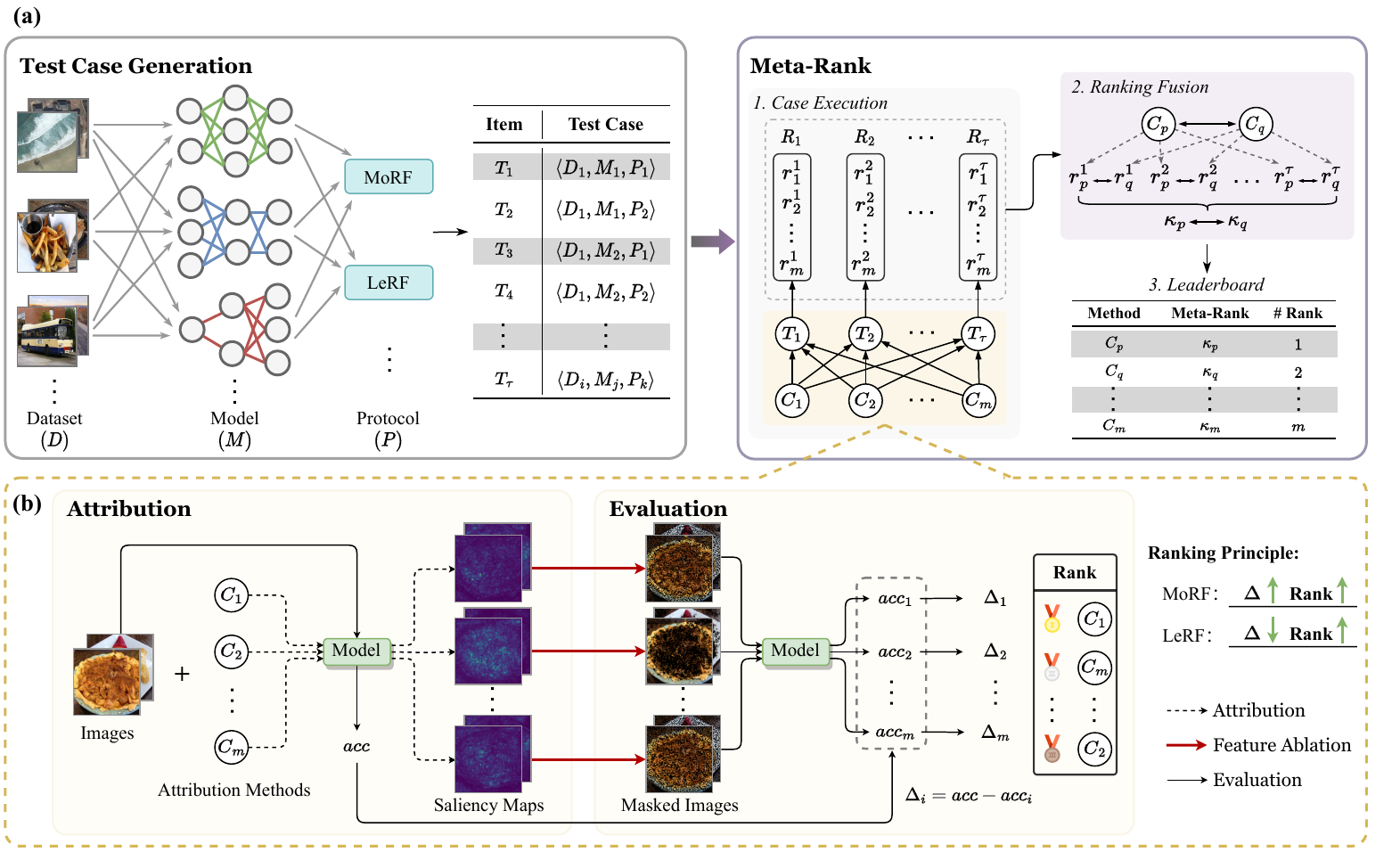}
\caption{Meta-Rank benchmark. It is mainly divided into two stages: \textbf{Test Case Generation} and \textbf{Meta-Rank}~(a).
Test Case Generation: multiple factors~(\ie, datasets, models and evaluation protocols) are combined to generate $\tau$ different cases.
Meta-Rank:
(1) Case Execution. All $m$ competitors~(\ie, attribution methods) are applied to these cases, resulting in a collection of rankings.
The details of attribution evaluation on an individual case are provided in~(b).
(2) Ranking Fusion. All rankings~\{$R_1$, $R_2$, $\ldots$, $R_\tau$\} are subsequently fed into this module. The comparison of two competitors is transformed into the differences in their rankings across all cases, then integrated and converted into the discrepancy in Meta-Ranks.
(3) Leaderboard. Ultimately, a unified leaderboard is obtained based on the Meta-Ranks of the competitors.
``$T$'' is the test case, ``$C$'' is the competitor, and ``$\kappa$'' is the Meta-Rank value.
}
\label{fig:Meta-Rank}
\end{figure*}

To acquire a comprehensive comprehension of the experimental configurations employed in previous investigations, we present a summary of the datasets and models utilized in these works, as depicted in Table~\ref{tab:settings}. Notably, despite the limited choice of datasets and models, there remains considerable variation in the evaluations conducted thus far. The issue of evaluation consistency has largely been overlooked in prior research endeavors.
To bridge this gap, in this section, we first establish standardized settings and subsequently provide a comprehensive outline of the Meta-Rank evaluation benchmark. The workflow of Meta-Rank is illustrated in Figure~\ref{fig:Meta-Rank}.

\subsection{Standardized Settings}
\label{sec:std-setting}
The main idea of the proposed Meta-Rank is evaluating attribution explanations on a diverse array of datasets, models, and evaluation protocols.

\noindent \textbf{Datasets.} To streamline standardized evaluations of attribution methods, we meticulously select four commonly used datasets for benchmarking: NWPU-RESISC45~\cite{cheng2017remote}, Food-101~\cite{bossard2014food}, ImageNet-1k~\cite{deng2009imagenet}, and Places-365~\cite{zhou2017places}. NWPU-RESISC45 is a medium-scale dataset for remote sensing image classification, containing 45 categories of remote sensing images covering various scenes such as streets, airports, factories, forests, grasslands, and deserts. Food-101 is a medium-scale dataset for fine-grained food image classification, compassing a total of 101 different food categories. ImageNet is a large-scale dataset employed for image classification, consisting of 1,000 different categories. Places-365 is a large-scale dataset designed for the purposes of scene recognition, covering a wide range of 365 diverse scene categories. These datasets allow us to examine each attribution method under various domains, providing us with a more thorough view of the current literature.
Further details are provided in~Appendix~C.

\noindent \textbf{Models.} The selection of models presents a prominent source of inconsistency in previous research. In order to comprehensively validate the effectiveness of the attribution techniques across various models, we adopt six widely employed convolutional neural networks (CNNs) to evaluate existing attribution methods. These CNNs include ResNet-18~\cite{he2016deep}, ResNet-50~\cite{he2016deep}, Inception-v4~\cite{szegedy2017inception}, VGG-19~\cite{simonyan2014very}, EfficientNet\_b2~\cite{tan2019efficientnet}, and DenseNet-121~\cite{huang2017densely}. It should be noted that most attribution methods are specifically tailored for CNNs and cannot be directly applied to vision transformers~(ViTs)~\cite{chefer2021transformer, ijcai2024protopformer, leem2024attention}. As such, for maintaining an equitable evaluation across attribution methods with a consistent set of evaluation settings, our benchmark does not take ViTs into consideration. Additional details regarding these used models can be found in Appendix C.

\noindent \textbf{Evaluation protocols.} Recent advances have discovered that different evaluation protocols, MoRF and LeRF~\cite{samek2016evaluating}, may produce conflicting rankings of attribution methods~\cite{rong2022consistent}. To provide a comprehensive view of the attribution methods, the proposed Meta-Rank considers both protocols for evaluation.

\noindent \textbf{Test case settings.}
To gain a generalized evaluation for attributions, our analytical experiments evaluate existing attribution methods on a wide array of experiment settings, necessitating the initial preparation of baselines.
We first train ResNet-18, Inception-v4 and VGG-19 on NWPU-RESISC45 and Food-101.
Subsequently, we exclusively train EfficientNet\_b2 and DenseNet-121 on the NWPU-RESISC45 dataset. These baselines are implemented with the PyTorch~\cite{paszke2019-pytorch} framework. Adam~\cite{kingma2014adam} is used to optimize the models, with the learning rate of $1 \times 10^{-4}$ and the weight decay rate of $5 \times 10^{-4}$. The cosine annealing is adopted as the learning rate decay schedule.
Furthermore, we also incorporate pre-trained ResNet-18, ResNet-50 sourced from Places-365, and pre-trained ResNet-18, Inception-v4, VGG-19 from ImageNet-1k.
The aforementioned baselines, combined with two evaluation protocols, are served as the test cases for the benchmark. Detailed information for each test case can be found in~Appendix~C.

\subsection{The Proposed Benchmark Metric}
\label{meta-rank}
We further propose a benchmark metric, named Meta-Rank, to facilitate a more comprehensive benchmark comparison between attribution methods. The main idea of the Meta-Rank metric is viewing each specific configuration of the model, the dataset and the evaluation protocol as one test case. Meta-Rank aggregates the ranking results~(\ie, rank the rankings) from a diverse array of test cases, thereby mitigating biases associated with specific configurations.

Formally, let the triplet $\langle D_i,M_j,P_k \rangle$ be one test case for evaluating the attribution algorithms, where $D$, $M$ and $P$ denote the dataset, model and evaluation protocol respectively. The subscript $i$, $j$ and $k$ denote one specific configuration from the candidate datasets, models and protocols as described in Section~\ref{sec:std-setting}. Let $\mathcal{C}=\left\{C_1, C_2, ..., C_m\right\}$ be the set of the competitors (\ie, attribution methods).
In the proposed benchmark, all the competitors are evaluated on every generated test case, thus producing a ranking among the competitors.
For the test case $\langle D_i,M_j,P_k \rangle$, we use $R_{(i,j,k)} = (r^{(i,j,k)}_1, r^{(i,j,k)}_2,...,r^{(i,j,k)}_m)$ to denote the rankings of all the competitors, where $r^{(i,j,k)}_m$ denotes the rank of the $m$-th attribution method. After evaluating the attribution methods on all the test cases, the probability that $C_p$ outperforms $C_q$ is formally defined as follows:
\begin{equation}
    \mathtt{P}_{q \prec p}=\frac{1}{|\Omega|} \sum\nolimits_{(i,j,k)}^{\Omega} \boldsymbol{1}(r^{(i,j,k)}_q \prec r^{(i,j,k)}_p) \text{,}
\end{equation}
% mathbbm
where $\boldsymbol{1}$ denotes the indicator function, and $r_q\prec r_p$ represents the ranking $r_q$ is outperformed by $r_p$. $\Omega$ is the joint space of datasets, models and evaluation protocols.
The probability $\mathtt{P}_{q \prec p}$ is then converted into the logistic odds, obtaining a more sensitive metric $\mathtt{Logit}_{q \prec p}$ that can better capture slight differences between $C_p$ and $C_q$:
\begin{equation}
    \mathtt{Logit}_{q \prec p} = \mathtt{logodds}(C_p, C_q)=\mathrm{log}(\mathtt{P}_{q \prec p} / (1-\mathtt{P}_{q \prec p})) \text{.}
\end{equation}
Based on the above derivation process, we can obtain a set of standardized results $\{\mathtt{Logit}_{q \prec p}\}$ that quantify the performance differences between any two competitors on all test cases. Then, we transform the performance discrepancy into the difference on the Meta-Rank metric:
\begin{equation}
\left\{
\begin{array}{ll}
    \kappa_1 - \kappa_2 = \mathtt{Logit}_{2 \prec 1} \text{,} & \text{\quad Competitor \#1 \vs \#2} \\
    \kappa_1 - \kappa_3 = \mathtt{Logit}_{3 \prec 1} \text{,} & \text{\quad Competitor \#1 \vs \#3} \\
    \kappa_1 - \kappa_4 = \mathtt{Logit}_{4 \prec 1} \text{,} & \text{\quad Competitor \#1 \vs \#4} \\
    \ldots & \quad \ldots \\
    \kappa_2 - \kappa_3 = \mathtt{Logit}_{3 \prec 2} \text{,} & \text{\quad Competitor \#2 \vs \#3} \\
    \kappa_2 - \kappa_4 = \mathtt{Logit}_{4 \prec 2} \text{,} & \text{\quad Competitor \#2 \vs \#4} \\
    \ldots & \quad \ldots \\
    \kappa_{m-1} - \kappa_m = \mathtt{Logit}_{m \prec m-1} \text{.} & \text{\quad Competitor \#m-1 \vs \#m}
\end{array}
\right.
\label{eq:meta-rank}
\end{equation}
Here, $\kappa_m$ represents the Meta-Rank value of $C_m$.
\begin{figure*}[!t]
\centering%
\subfloat[Food-ResNet18(M)]{%
    \includegraphics[width=0.25\linewidth]{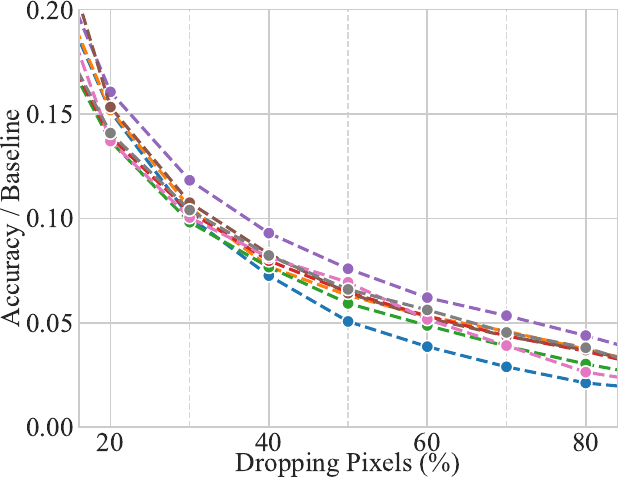}%
    \label{food-res}%
}%
\subfloat[Food-ResNet18(L)]{%
    \includegraphics[width=0.25\linewidth]{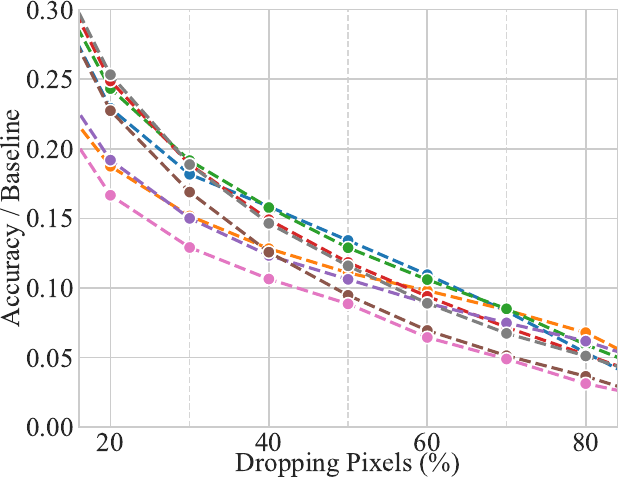}%
    \label{food-res-reverse}%
}%
\subfloat[IN-ResNet18(M)]{%
    \includegraphics[width=0.25\linewidth]{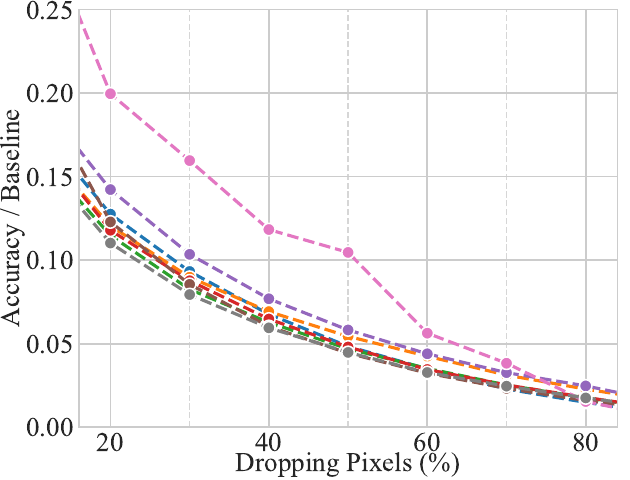}%
    \label{image-res}%
}%
\subfloat[IN-ResNet18(L)]{
    \includegraphics[width=0.25\linewidth]{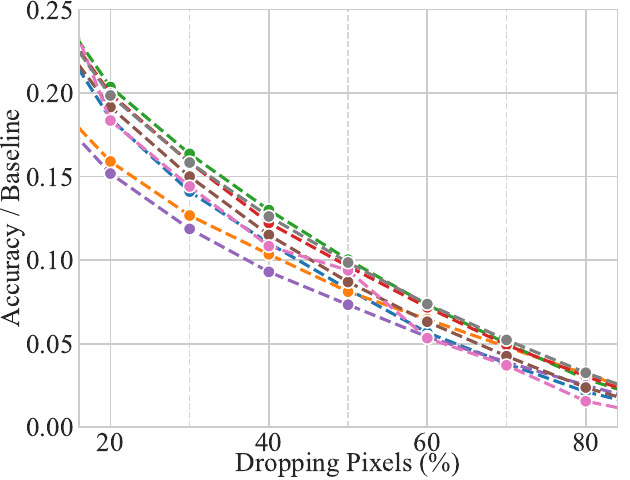}%
    \label{image-res-reverse}%
}%
\\
\subfloat[Food-Inceptionv4(M)]{
    \includegraphics[width=0.25\linewidth]{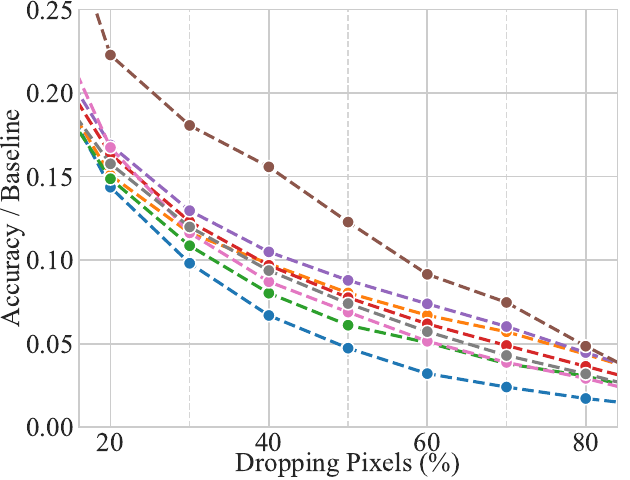}%
    \label{food-inception}%
}%
\subfloat[Food-Inceptionv4(L)]{%
    \includegraphics[width=0.25\linewidth]{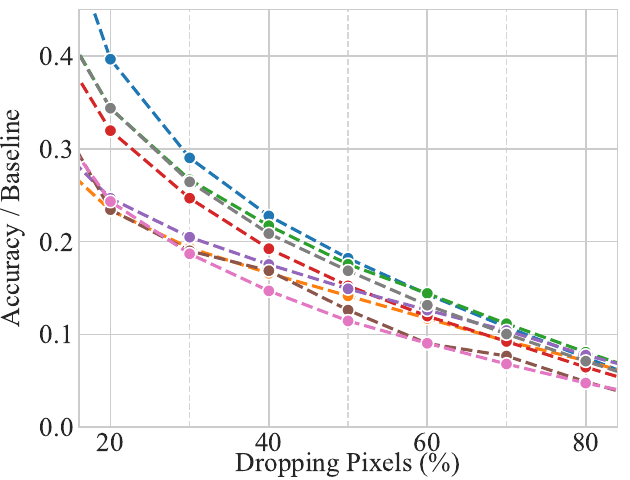}%
    \label{food-inception-reverse}%
}%
\subfloat[IN-Inceptionv4(M)]{%
    \includegraphics[width=0.25\linewidth]{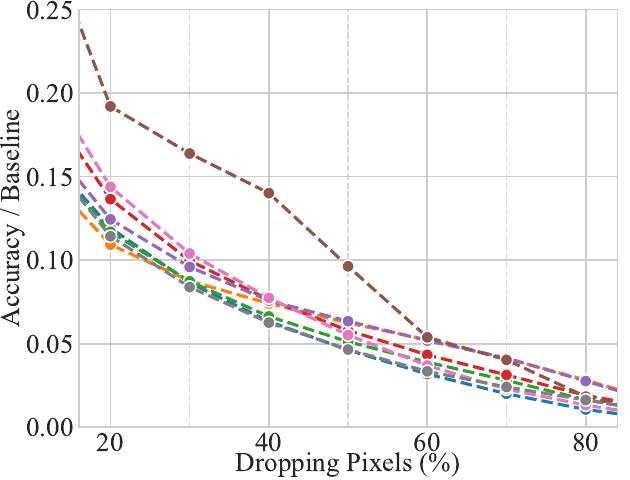}%
    \label{image-inception}%
}%
\subfloat[IN-Inceptionv4(L)]{%
    \includegraphics[width=0.25\linewidth]{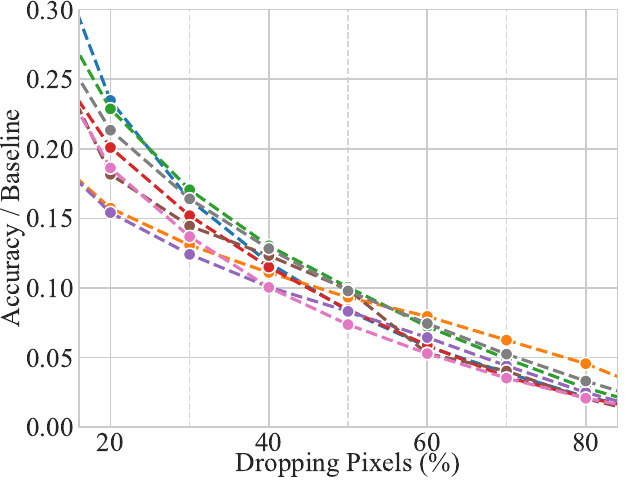}%
    \label{image-inception-reverse}%
}%
\\
\subfloat[Food-VGG19(M)]{%
    \includegraphics[width=0.25\linewidth]{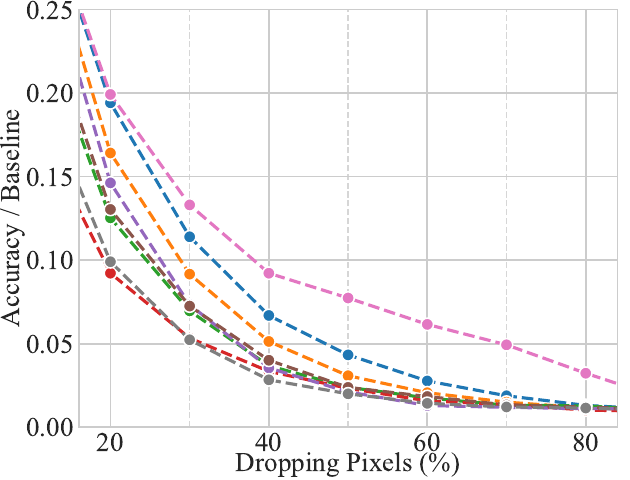}%
    \label{food-vgg}%
}%
\subfloat[Food-VGG19(L)]{%
    \includegraphics[width=0.25\linewidth]{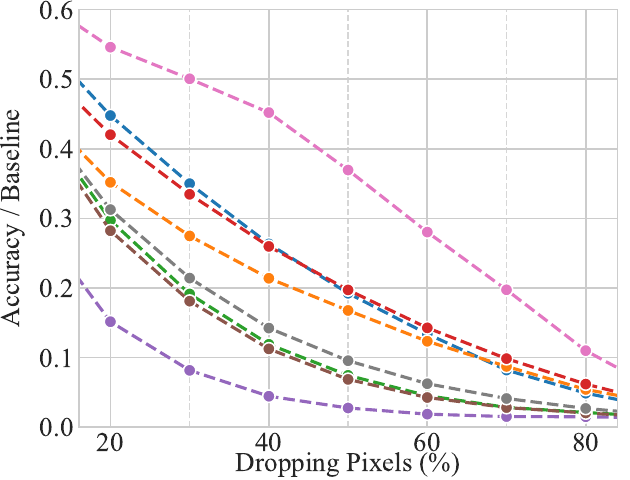}%
    \label{food-vgg-reverse}%
}%
\subfloat[IN-VGG19(M)]{%
    \includegraphics[width=0.25\linewidth]{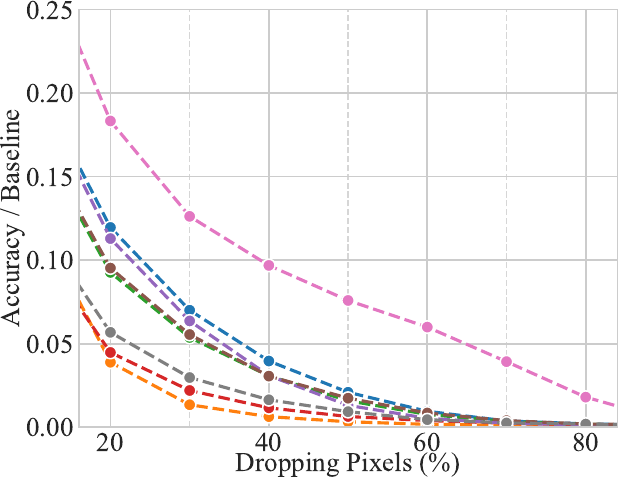}%
    \label{image-vgg}%
}%
\subfloat[IN-VGG19(L)]{
    \includegraphics[width=0.25\linewidth]{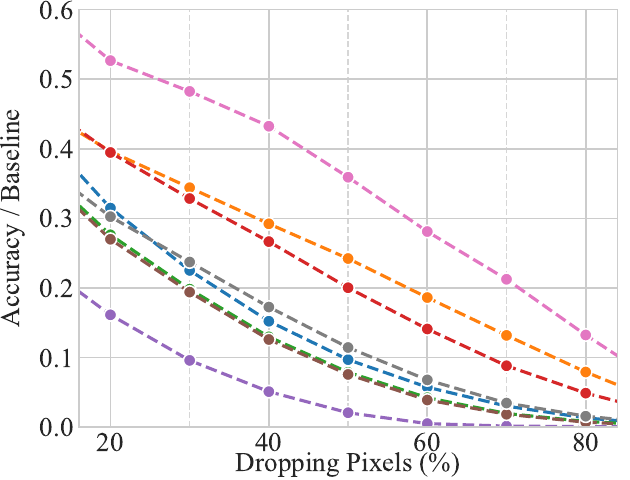}%
    \label{image-vgg-reverse}%
}%
\\
\includegraphics[width=0.98\linewidth]{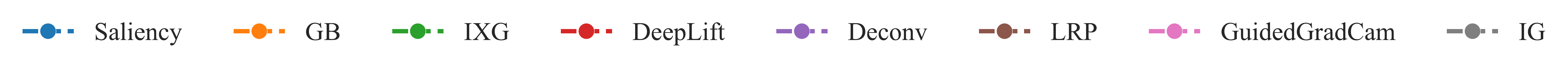}
\caption{Evaluation results on 12 test cases. These cases consist of two datasets: Food-101~(a)(b)(e)(f)(i)(j), ImageNet-1k~(c)(d)(g)(h)(k)(l), three models: ResNet-18~(a)(b)(c)(d), Inception-v4~(e)(f)(g)(h), VGG-19~(i)(j)(k)(l), and two protocols: MoRF(a)(c)(e)(g)(i)(k), LeRF(b)(d)(f)(h)(j)(l).
``Baseline'' represents the accuracy of the model when no features are ablated.
}
\label{fig:inconsis}
\end{figure*}
After solving the system of equations in~Equation~\ref{eq:meta-rank}, we can derive a score $\kappa$ for each competitor, serving as an identifier for its comprehensive performance. Through the arrangement of competitors in descending order of their $\kappa$ values, a unified leaderboard can finally be constructed.

\section{Main Results}
This section provides a detailed elaboration of extensive experiments on Meta-Rank, with a particular emphasis on the following pivotal questions.
\begin{enumerate}[label=\textbullet]
    \item \textbf{RQ1:} Are attribution methods consistently effective across diverse settings~(\ie, datasets, models, evaluation protocols) and distinct checkpoints?~(Section~\ref{sec:inconsis})
    \item \textbf{RQ2:} Does the state-of-the-art evaluation method, ROAD, yield consistent rankings across different test cases?
    (Section~\ref{sec:road})
    \item \textbf{RQ3:} According to our proposed Meta-Rank, which attribution methods exhibit superior performance across a wide range of cases?~(Section~\ref{sec:meta-eval})
    \item \textbf{RQ4:} Can Meta-Rank conduct the evaluation of substantial methods within a feasible timeframe? ~(Section~\ref{sec:time})
\end{enumerate}

\begin{figure*}[!t]
\centering%
\subfloat[MoRF]{%
    \includegraphics[width=0.34\linewidth]{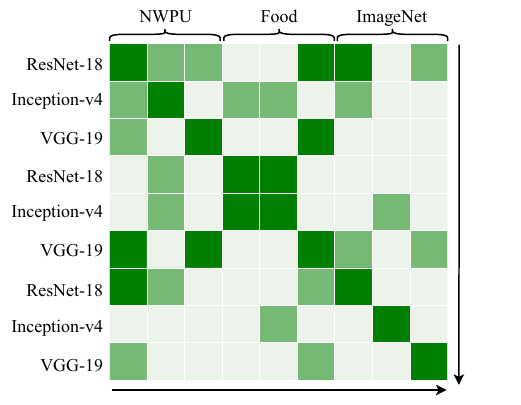}%
    \label{Spearman-morf}%
}%
\subfloat[LeRF]{%
    \includegraphics[width=0.34\linewidth]{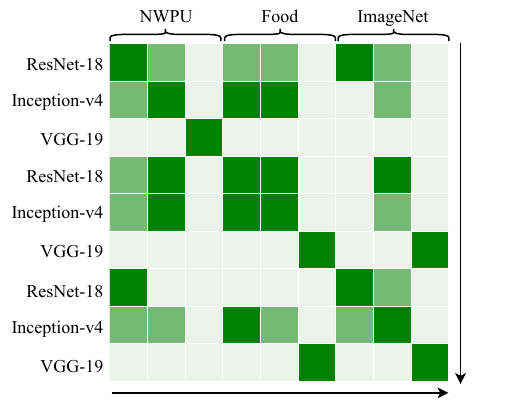}%
    \label{Spearman-lerf}%
}%
\subfloat[MoRF $\textit{v.s.}$ LeRF]{%
    \includegraphics[width=0.3\linewidth]{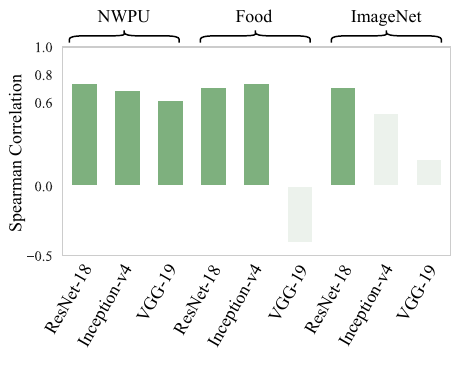}%
    \label{Spearman-ml}%
}%
\caption{
Spearman correlation among nine test cases~(NWPU, Food, and ImageNet datasets paired with ResNet-18, Inception-v4, and VGG-19 models) in MoRF~(a) and LeRF~(b), and between MoRF and LeRF on the same nine cases~(c).
The labels along the horizontal arrow are equivalent to those along the vertical arrow.
Here, we define a correlation score higher than 0.8 as strongly correlated~(\textcolor{green1}{\rule{0.66em}{0.66em}}), between 0.6 and 0.8 as moderately correlated~(\textcolor{green2}{\rule{0.66em}{0.66em}}), and lower than 0.6 as weakly correlated~(\textcolor{green3}{\rule{0.66em}{0.66em}}).}
\label{fig:Spearman-corr}
\end{figure*}

\subsection{RQ1: Consistency Investigation}
\label{sec:inconsis}
In this section, we investigate the consistency of performance rankings for existing attribution methods across diverse evaluation settings. Specifically, we focus on the model-, dataset-, protocol- and checkpoint-level ranking consistencies by varying these factors in the evaluation.
To this end, we assess the aforementioned eight attribution methods on all the generated test cases, \ie, with six models, four datasets and two ablation protocols (MoRF and LeRF). This yields a total of 26 groups of results, each corresponding to one test case. Some experimental results are presented in Figure~\ref{fig:inconsis}.
For more details of the experimental results, please refer to~Appendix~D.

\noindent{\textbf{Model-level Consistency.}}
It is evident that the attribution methods exhibit varying performance across different models.
To elaborate:
(1) LRP has the worst performance on the ``ImageNet-Inceptionv4(M)'' case but significantly better results on the ``ImageNet-ResNet18(M)'' and ``ImageNet-VGG19(M)'' cases.
(2) Integrated Gradients and Guided Grad-CAM exhibit comparable performance on the ``Food-ResNet18(M)'' case, but show great disparity on the ``Food-VGG19(M)'' case.
For a more clear measurement of the inconsistency, we calculate the \textit{Spearman Rank correlations} among several cases and illustrate them in Figure~\ref{fig:Spearman-corr}.
Here, we extract \textit{area under the curve}~(AUC)~\cite{atanasova2020diagnostic} of all attribution methods~(detailed in Appendix~D) to generate a ranking list for each case.
In the MoRF protocol, removing the most critical pixels identified by the attribution method causes a substantial drop in model performance. Thus, a lower AUC value indicates a superior method. And the LeRF protocol follows the inverse principle.
The results suggest a pronounced weakness in the consistency at the model level, with a strong correlation found in only $2$ cases~(\ie, the correlation between ResNet-18 and Inception-v4 on the Food-101 dataset in both MoRF and LeRF protocols) out of $18$ cases.

\noindent{\textbf{Dataset-level Consistency.}}
Analogously, attribution methods have shown noticeable inconsistency in performance on different datasets. For instance, Guided Backpropagation demonstrates superior performance on ``ImageNet-VGG19(M)'' yet ranks third-worst on ``Food-VGG19(M)''. In addition, according to the results of Spearman analysis, only $\frac{5}{18}$ cases show strong correlations at the dataset level.

\noindent{\textbf{Protocol-level Consistency.}}
The evaluation results indicate that significant inconsistencies between the MoRF and LeRF protocols still exist.
A detailed comparison between ``Food-Inceptionv4(M)'' and ```Food-Inceptionv4(L)'' reveals marked discrepancies in Saliency method under MoRF versus LeRF. Besides, the Spearman correlation between MoRF and LeRF is computed for nine test cases~(in Figure~\ref{fig:Spearman-corr}~(c)). It can be easily seen that the correlation of all cases at the protocol level is either moderately correlated or weakly correlated, with no strong correlation observed.

\begin{table}[!t]
    \centering
    \begin{minipage}[!t]{0.48\textwidth}
        \centering%
        \caption{AUC rankings of the eight methods at different checkpoints along the same trajectory. Here, ResNet-18 trained on NWPU-RESISC45 is taken as an illustrative example.}%
        \includegraphics[width=0.98\linewidth]{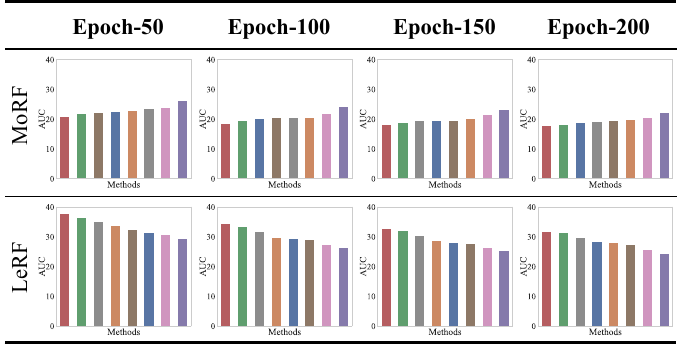}%
        \\
        \includegraphics[width=0.98\linewidth]{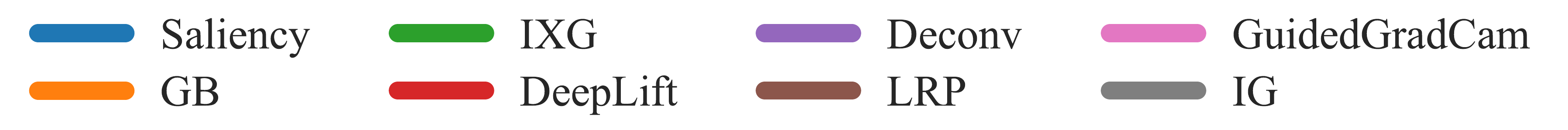}
        \label{fig:consis-epoch}
    \end{minipage}
    \quad
    \begin{minipage}[!t]{0.484\textwidth}
    \tiny
        \centering%
        \caption{Comparison of Spearman results among four cases between ROAD and feature ablation. $a \mid b$ represents a pair of Spearman results, where $a$ denotes using feature ablation~(in Section~\ref{sec:inconsis}) and $b$ denotes using noisy linear imputation.
        The settings of color blocks in this table are the same as those in Figure~\ref{fig:Spearman-corr}.
        \label{tab:spearman-road}%
        }
        \begin{tabular}{*{1}{l} *{4}{c}}
        \toprule
        \multicolumn{1}{l}{\multirow{2}{*}{\textbf{}}}
        & \multicolumn{2}{c}{\textbf{NWPU-ResNet18}}
        & \multicolumn{2}{c}{\textbf{Food-Inceptionv4}}\\
        \cmidrule(lr){2-3} \cmidrule(lr){4-5}
        & \multicolumn{1}{c}{MoRF~(\#1)}
        & \multicolumn{1}{c}{LeRF~(\#2)}
        & \multicolumn{1}{c}{MoRF~(\#3)}
        & \multicolumn{1}{c}{LeRF~(\#4)} \\
        \midrule
    	\#1 & \textcolor{green1}{\rule{1.2em}{1.2em}} $\mid$ \textcolor{green1}{\rule{1.2em}{1.2em}} & \textcolor{green2}{\rule{1.2em}{1.2em}} $\mid$ \textcolor{green2}{\rule{1.2em}{1.2em}} & \textcolor{green3}{\rule{1.2em}{1.2em}} $\mid$ \textcolor{green2}{\rule{1.2em}{1.2em}} & \textcolor{green3}{\rule{1.2em}{1.2em}} $\mid$ \textcolor{green3}{\rule{1.2em}{1.2em}} \\
        \#2 & \textcolor{green2}{\rule{1.2em}{1.2em}} $\mid$ \textcolor{green2}{\rule{1.2em}{1.2em}} & \textcolor{green1}{\rule{1.2em}{1.2em}} $\mid$ \textcolor{green1}{\rule{1.2em}{1.2em}} & \textcolor{green3}{\rule{1.2em}{1.2em}} $\mid$ \textcolor{green2}{\rule{1.2em}{1.2em}} & \textcolor{green2}{\rule{1.2em}{1.2em}} $\mid$ \textcolor{green3}{\rule{1.2em}{1.2em}} \\
        \#3 & \textcolor{green3}{\rule{1.2em}{1.2em}} $\mid$ \textcolor{green2}{\rule{1.2em}{1.2em}} & \textcolor{green3}{\rule{1.2em}{1.2em}} $\mid$ \textcolor{green2}{\rule{1.2em}{1.2em}} & \textcolor{green1}{\rule{1.2em}{1.2em}} $\mid$ \textcolor{green1}{\rule{1.2em}{1.2em}} & \textcolor{green2}{\rule{1.2em}{1.2em}} $\mid$ \textcolor{green3}{\rule{1.2em}{1.2em}} \\
        \#4 & \textcolor{green3}{\rule{1.2em}{1.2em}} $\mid$ \textcolor{green3}{\rule{1.2em}{1.2em}} & \textcolor{green2}{\rule{1.2em}{1.2em}} $\mid$ \textcolor{green3}{\rule{1.2em}{1.2em}} & \textcolor{green2}{\rule{1.2em}{1.2em}} $\mid$ \textcolor{green3}{\rule{1.2em}{1.2em}} & \textcolor{green1}{\rule{1.2em}{1.2em}} $\mid$ \textcolor{green1}{\rule{1.2em}{1.2em}} \\
        \bottomrule
        \end{tabular}
    \end{minipage}
\end{table}

\noindent{\textbf{Checkpoint-level Consistency.}}
We also consider the consistency of attribution methods at different training stages of the model.
To illustrate, using ResNet-18 trained on NWPU-RESISC45 as a case study, we extract distinct checkpoints from the same training trajectory and employ two evaluation protocols for attribution analysis.
The results of the AUC rankings are recorded in Table~\ref{fig:consis-epoch}. It can be easily observed that the rankings of attribution methods demonstrate high consistency across different checkpoints.
In MoRF, the outcomes of Epoch-200 differ by only one distinction from those at both Epoch-100 and Epoch-150, while exhibiting slightly greater disparities when compared to those at Epoch-50. Similarly for LeRF, the rankings at Epoch-200 also exhibit a single deviation from those at Epoch-100 and Epoch-150, respectively, and show two distinctions in comparison to those at Epoch-50. These findings suggest a notable consistency in the ranking of attribution methods across different checkpoints. Moreover, the rankings tend to stabilize rapidly as the model converges.
More detailed results under other settings can be found in Appendix~D.

\subsection{RQ2: Necessity of Meta-Rank}
\label{sec:road}
To further substantiate the indispensability of the Meta-Rank benchmark, we employ the state-of-the-art attribution evaluation method known as ROAD for assessing the attribution methods across various cases.
Starting from information theoretic analyses, ROAD identifies the class information leakage problem dictated by the ablation mask and finally derives the \textit{noisy linear imputation} for minimally revealing imputation, which is empirically demonstrated to produce consistent rankings of attribution methods between evaluations with MoRF and LeRF. However, it overlooks consistency at both the model and dataset levels.
Thus, we randomly select four cases, namely NWPU-ResNet18(M), NWPU-ResNet18(L), Food-Inceptionv4(M), and Food-Inceptionv4(L), to assess the eight aforementioned attribution methods using ROAD. The evaluation metric used in this study aligns with those in RQ1. Table~\ref{tab:spearman-road} presents the Spearman correlations among the evaluation results of the four cases. Further information regarding the accuracy decay curves is available in Appendix~E.

It can be easily seen that, even when ROAD is employed as a substitute for feature ablation, the correlation between the rankings of attribution methods on different cases remains inconsistent. Specifically, all correlations demonstrate either moderate or weak, with no instances of strong correlation.
This indicates that relying solely on ROAD for evaluating attribution methods across distinct cases is inadequate.
Consequently, there is a critical need for a cross-case benchmark to ensure robust ranking results.

\begin{table}[!t]
    \centering
    \begin{minipage}[!t]{0.488\textwidth}
        \tiny% 
        \centering%
        \caption{Leaderboard of the eight attribution methods. The larger Meta-Rank indicates the stronger ability of attribution method.
        The top three methods exhibit more positive performance, while the remaining five methods perform negatively.
        }
        \begin{tabular}{l|r|c}
        \toprule
    	\textbf{Methods} & \textbf{Meta-Rank Score} & \textbf{ \#Rank } \\
        \midrule
        \rowcolor{gray!18}
    	Input$\odot$Gradient & $2.3200$ & $1$ \\
        \rowcolor{gray!18}
    	Integrated Gradients & $2.0736$ & $2$ \\
        \rowcolor{gray!18}
    	DeepLift & $1.0780$ & $3$ \\
    	Saliency & $-0.0483$ & $4$ \\
    	LRP & $-0.2234$ & $5$ \\
    	Guided Backpropagation & $-0.2585$ & $6$ \\
    	Guided Grad-CAM & $-0.9679$ & $7$ \\
    	Deconvolution & $-3.9734$ & $8$ \\
        \bottomrule
        \end{tabular}
        \label{tab:Meta-Rank}
    \end{minipage}
    \quad
    \begin{minipage}[!t]{0.476\textwidth}
    \centering
        \includegraphics[width=0.88\linewidth]{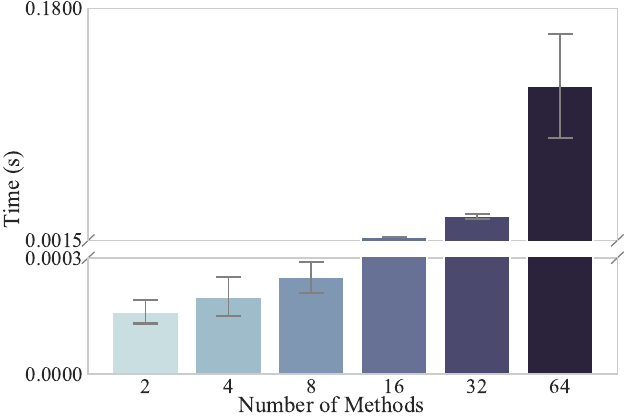}
        \captionof{figure}{Time consumption of the ``Ranking Fusion'' module in Meta-Rank. $2$, $4$, $8$, $16$, $32$, and $64$ are the number of evaluated attribution methods. The gray bar represents the relative error.}
        \label{fig:time}
    \end{minipage}
\end{table}

\subsection{RQ3: Attribution Evaluation with Meta-Rank}
\label{sec:meta-eval}
Following the evaluation scheme outlined in the Meta-Rank benchmark~(described in Section~\ref{sec:meta}), we have carried out a unified measurement of the eight attribution methods.
The consistency validation experiments in Section~\ref{sec:inconsis} generate evaluation results for attribution methods across distinct configurations, encompassing four datasets, six models, and two protocols, resulting in 26 sets of data. We then transform these results into AUC scores, which serve as the performance metric for the attribution methods.

Meta-Rank generates a comprehensive value for each attribution method.
These uniquely determined scores enable us to establish a unified ranking list for all the methods.
The finalized leaderboard for the eight attribution methods is presented in Table~\ref{tab:Meta-Rank}.
The results reveal Input$\odot$Gradient as the top-performing approach, while Deconvolution exhibits the poorest performance.
(1) In particular, Input$\odot$Gradient achieves the highest score among the eight methods.
By element-wise multiplying the gradients with the inputs, Input$\odot$Gradient produces more stable attributions in the face of perturbations, which contributes to its superior performance across multiple test cases.
(2) Furthermore, the second-ranked method, Integrated Gradients, also shows great potential, with its score closely approaching that of the top-ranked method.
(3) In contrast, Deconvolution obtains the lowest score, likely attributable to the hierarchical structure of the network accumulating noise during backpropagation.
It is worth noting that a Meta-Rank score can assume both positive and negative values. Positive values signify comparatively superior attribution performance, whereas negative values indicate inferior performance.

The experimental findings demonstrate that Meta-Rank furnishes a comprehensive benchmark for attribution evaluation across diverse test cases, while effectively ameliorating the issue of inconsistent performance across heterogeneous cases.
It computes the discrepancies between methods and converts them into the differences in capacity to derive the Meta-Rank.
This result is unique, consistent, and aligned with practical application needs.

\subsection{RQ4: Efficiency of Meta-Rank}
\label{sec:time}
To measure the time consumption, we conduct experiments on the CPU device~(Intel(R) Xeon(R) Platinum 8260L CPU @ 2.40GHz) using different numbers of methods. Here, we focus on the additional ``Ranking Fusion'' module, as the ``Case Execution'' module has been generically applied in all evaluations. In particular, we compare the time expenses of Meta-Rank during the calculation of the $2$, $4$, $8$, $16$, $32$, and $64$ methods. It is important to highlight that the first three test groups~(\ie, $2$, $4$, and $8$ methods) rely on real data gathered from previous experiments, whereas the last three test groups~(\ie, $16$, $32$, and $64$ methods) employ simulated data randomly generated. The data here refers to a set of accuracies~(\ie, $\{acc_i\}$ in Figure~\ref{fig:Meta-Rank}) obtained on each test case according to every attribution method.

The results in Figure~\ref{fig:time} indicate that as the number of attribution methods increases, the computation time also increases. However, since the number of attribution methods is finite, the computational time remains within an acceptable range. For example, when $64$ methods are involved in the Meta-Rank evaluation, the time required to generate the leaderboard is approximately $0.12$s, which completely meets expectations. Additionally, using GPUs for Meta-Rank calculations can further enhance the computation speed. This efficiency is beneficial for the promotion and application of Meta-Rank.

\section{Discussion}
\noindent \textbf{Choice of faithfulness metric.} Although there are many dimensions for attribution evaluation in previous work~\cite{hedstrom2023quantus}, faithfulness remains the most solid and general criteria for benchmarking attribution methods. In the context of multiple experimental settings, as such, we opt to use it as the base metric $\Delta$ in our framework to facilitate subsequent Meta-Rank assessment and then enable broader evaluation on novel tasks. Meanwhile, Meta-Rank can also accommodate evaluations on other dimensions by enabling the derivation of a ranking list along each respective evaluation metric.

\noindent \textbf{Missingness bias in feature ablation.}
Previous work has discussed the missingness bias problem caused by the removal of pixels through masking values in CNNs~\cite{jain2022missingness}. Hooker~\etal~proposed ROAR to address this issue, but it inadvertently introduced the problem of information leakage. Subsequently, Rong~\etal~developed ROAD to mitigate this leakage, but it failed to fully remove the pixels~(explained in~Appendix E).
We recognize that completely eliminating this bias poses substantial difficulties, since the root of missingness bias lies in the operation of removing pixels within CNNs.
Furthermore, the impact of bias exhibits variability across individual cases, also posing additional challenges in quantification and measurement.
However, we believe that with a more diverse array of test cases, the missingness bias is more neutralized.
Future research will further explore solutions for eliminating this bias.

\noindent \textbf{Scalability of Meta-Rank.}
When introducing new attribution methods to the ranking, we initially conduct tests on several existing cases. The time complexity of evaluating a method on an individual case is linear, which has been demonstrated in previous studies~\cite{petsiuk2018rise, rong2022consistent}.
Subsequently, the obtained results, in conjunction with historical data, are fed into the ``Ranking Fusion'' module to generate an updated ranking list. The time overhead associated with this process is also optimistic, as illustrated in Section~\ref{sec:time}.
This flexibility allows for the inclusion of any attribution method into the leaderboard at any time.
Future endeavors will focus on expanding the corpus of test cases and pursuing further computational efficiency by leveraging parallel computing paradigms.

\section{Conclusion}
To address the lack of systematic research in feature attribution, we propose Meta-Rank, a novel standardized benchmark for robust and unified evaluation of methods. Meta-Rank aggregates relative comparisons between all method pairs across diverse settings, mitigating inconsistencies.
Through extensive experiments, we reveal three critical insights: diverse evaluation settings yield divergent performance rankings, assessment results remain consistent across distinct checkpoints along the same training trajectory, and previous attempts at consistent evaluation exhibit limitations on heterogeneous models and datasets.
Meta-Rank provides a leaderboard that highlights the broader applicability of Input$\odot$Gradient and Integrated Gradients.
Remarkably, Meta-Rank exhibits exceptional computational efficiency, enabling ranking computations to be completed within milliseconds even for a substantial number of methods.
In the future, we hope the proposed benchmark will facilitate the development of faithful and consistent explanations, fostering transparency and accountability.

\section*{Acknowledgements}
This work is supported by National Natural Science Foundation of China~(62106220), Alibaba-Zhejiang University Joint Research Institute of Frontier Technologies, Ningbo 2025 Science and Technology Innovation Major Project (No. 2022Z072), and the Fundamental Research Funds for the Central Universities~(226-2024-00058).

% \clearpage  % TODO REVIEW/FINAL: This \clearpage needs to be removed from both review and camera-ready versions.

\bibliographystyle{splncs04}
\bibliography{main}

\end{document}